\documentclass[journal]{IEEEtran}

\usepackage{xurl}

\usepackage{fontspec}
\usepackage{polyglossia}

\newfontfamily\pashtofont[
  Script=Arabic,
  Scale=1.1
]{FreeSerif.ttf}[BoldFont=FreeSerifBold.ttf, BoldItalicFont=FreeSerifBoldItalic.ttf]

\setdefaultlanguage{english}
\setotherlanguage{arabic}
\newfontfamily\arabicfont[Script=Arabic,Scale=1.1]{FreeSerif.ttf}[BoldFont=FreeSerifBold.ttf, BoldItalicFont=FreeSerifBoldItalic.ttf]

\usepackage{newunicodechar}
\newunicodechar{ټ}{{\pashtofont ټ}}  
\newunicodechar{ډ}{{\pashtofont ډ}}  
\newunicodechar{ڼ}{{\pashtofont ڼ}}  
\newunicodechar{ړ}{{\pashtofont ړ}}  
\newunicodechar{ښ}{{\pashtofont ښ}}  
\newunicodechar{ږ}{{\pashtofont ږ}}  
\newunicodechar{ځ}{{\pashtofont ځ}}  
\newunicodechar{څ}{{\pashtofont څ}}  
\newunicodechar{ۍ}{{\pashtofont ۍ}}  
\newunicodechar{ئ}{{\pashtofont ئ}}  
\newunicodechar{ت}{{\pashtofont ت}}
\newunicodechar{د}{{\pashtofont د}}
\newunicodechar{ن}{{\pashtofont ن}}
\newunicodechar{ج}{{\pashtofont ج}}
\newunicodechar{ش}{{\pashtofont ش}}
\newunicodechar{خ}{{\pashtofont خ}}
\newunicodechar{ر}{{\pashtofont ر}}
\newunicodechar{ي}{{\pashtofont ي}}
\newunicodechar{ل}{{\pashtofont ل}}
\newunicodechar{ط}{{\pashtofont ط}}
\newunicodechar{م}{{\pashtofont م}}
\newunicodechar{ک}{{\pashtofont ک}}
\newunicodechar{ګ}{{\pashtofont ګ}}
\newunicodechar{ز}{{\pashtofont ز}}
\newunicodechar{س}{{\pashtofont س}}
\newunicodechar{ص}{{\pashtofont ص}}
\newunicodechar{غ}{{\pashtofont غ}}
\newunicodechar{ق}{{\pashtofont ق}}
\newunicodechar{ې}{{\pashtofont ې}}  
\newunicodechar{ۀ}{{\pashtofont ۀ}}  
\newunicodechar{ا}{{\pashtofont ا}}
\newunicodechar{ب}{{\pashtofont ب}}
\newunicodechar{پ}{{\pashtofont پ}}
\newunicodechar{و}{{\pashtofont و}}
\newunicodechar{ه}{{\pashtofont ه}}
\newunicodechar{ی}{{\pashtofont ی}}  
\newunicodechar{ف}{{\pashtofont ف}}
\newunicodechar{ع}{{\pashtofont ع}}
\newunicodechar{آ}{{\pashtofont آ}}
\newunicodechar{ح}{{\pashtofont ح}}

\newcommand{\ps}[1]{{\pashtofont\RL{#1}}}

\usepackage{graphicx}
\usepackage{booktabs}
\usepackage{tabularx}
\usepackage{multirow}
\usepackage{array}
\usepackage{amsmath}
\usepackage{placeins}        
\providecommand{\href}[2]{#2}

\begin{document}

\title{Benchmarking Multilingual Speech Models on Pashto:\\
Zero-Shot ASR, Script Failure, and Cross-Domain Evaluation}

\author{Hanif~Rahman,~\IEEEmembership{Independent Researcher}}

\markboth{IEEE/ACM Transactions on Audio, Speech, and Language Processing}%
{Rahman: Benchmarking Multilingual Speech Models on Pashto}

\maketitle

\begin{abstract}
Pashto is spoken by approximately 60--80 million people but has no published
benchmarks for multilingual automatic speech recognition (ASR) on any shared
public test set.
This paper reports the first reproducible multi-model evaluation on public
Pashto data, covering zero-shot ASR, script-level failure, and cross-domain
evaluation of fine-tuned models.
For zero-shot ASR, ten models (all seven Whisper sizes, MMS-1B,
SeamlessM4T-v2-large, and OmniASR-CTC-300M) are evaluated on the FLEURS
Pashto test set and a filtered Common Voice~24 subset; zero-shot Whisper WER
ranges from 90\% to 297\%, with the medium model collapsing to 461\% on
Common Voice~24 consistent with decoder looping.
SeamlessM4T achieves 39.7\% WER on Common Voice~24 (the best zero-shot
result reported to date, as of submission); MMS-1B achieves 43.8\% on FLEURS.
For script failure, a language-identification audit shows that no Whisper
model produces Pashto-script output in more than 0.8\% of utterances, while
MMS-1B, SeamlessM4T, and OmniASR each exceed 93\% Pashto-script fidelity;
WER alone does not reveal this failure, since a model generating Arabic-script
output on Pashto audio has not achieved ASR in any interpretable sense.
For cross-domain evaluation, five fine-tuned Pashto ASR models are evaluated
on both test sets: published WER figures of 14\% degrade to 32.5--59\% on
out-of-distribution sets, while one augmented model achieves 35.1\% on both
sets with zero cross-domain degradation.
Character-class error stratification confirms that Pashto-unique phonemes
(the retroflex series and lateral fricatives) account for disproportionate
error mass.
All evaluations cover read speech only.
Five structural impediments to cumulative progress are identified and five
ordered research priorities are argued.

\end{abstract}


\section{Introduction}
\label{sec:introduction}

Pashto is spoken as a first language by approximately 60--80 million people,
comparable in speaker population to Korean, Italian, and Ukrainian---languages
for which commercial ASR and TTS have been available for over a decade.
No published benchmark exists for multilingual ASR on Pashto using any
shared public test set, and no open-source TTS system exists.
All results in this paper apply to read speech; no spontaneous or dialectal
Pashto evaluation data exists.

The only prior result on a public test set is 14\% WER from a fine-tuned
system~\cite{PashtoWhisper2024}; no zero-shot baseline contextualising that
figure has been reported.
Sehar et al.~\cite{Sehar2025Whisper} evaluate zero-shot Whisper on a private broadcast
corpus, but apply a non-deterministic GPT normalisation step, preventing
direct comparison.
This paper fills three specific gaps.

\textbf{Contribution 1: Zero-shot ASR benchmarks.}
We evaluate all seven Whisper sizes~\cite{Radford2023Whisper}, MMS-1B~\cite{Pratap2023MMS},
SeamlessM4T-v2-large~\cite{SeamlessCommunication2023},
and OmniASR-CTC-300M~\cite{OmniASR2025} on the FLEURS Pashto test
set~\cite{Conneau2022FLEURS} and on the full Common Voice 24 filtered test
set~\cite{Ardila2020CommonVoice}, with a fully deterministic evaluation pipeline
(Section~\ref{sec:protocol}).
SeamlessM4T achieves 39.7\% WER on CV24\_filtered (the best zero-shot result
reported to date, as of submission), 22 pp below MMS-1B on the same set.

\textbf{Contribution 2: Script failure as a distinct failure mode.}
A language-identification audit shows that no Whisper model produces
Pashto-script output in more than 0.8\% of utterances, while MMS-1B,
SeamlessM4T, and OmniASR each exceed 93\% Pashto-script fidelity.
WER alone does not distinguish script failure from acoustic mismatch;
reporting script fidelity is necessary for valid cross-model comparison.

\textbf{Contribution 3: Cross-domain evaluation of fine-tuned models.}
Five fine-tuned Pashto ASR models across three architectures and two training
corpora are evaluated on both test sets.
The best published 14\% WER becomes 59\% on FLEURS and 35\% on CV24\_filtered.
Data augmentation eliminates cross-domain degradation: \texttt{w2v-b2-aug}
achieves 35.1\% WER on both sets with zero degradation.
Character-class WER stratification confirms that Pashto-unique phonemes
(the retroflex series and lateral fricatives) account for disproportionate
error mass.

Without shared benchmarks a field cannot measure whether its systems improve.
This paper provides the reference points needed for that measurement.

\section{Phonological and Orthographic Background}
\label{sec:background}

\subsection{Pashto-Unique Phonemes}
\label{sec:phonemes}

Pashto is an Eastern Iranian language whose phoneme inventory differs from every
language well-represented in published multilingual speech model training corpora.
Four properties are directly relevant to the results below.

Pashto has a four-member retroflex stop series: voiceless \textsc{ټ}~(/ʈ/),
voiced \textsc{ډ}~(/ɖ/), retroflex nasal \textsc{ڼ}~(/ɳ/), and retroflex
flap \textsc{ړ}~(/ɽ/).
Each is phonemically contrastive with its dental counterpart---a contrast
absent from Arabic, Dari, and Urdu.

Pashto also has two phonemes absent from virtually all other multilingual ASR
training corpora: \textsc{ښ} and \textsc{ږ}.
In the Kandahari written standard, \textsc{ښ} is a voiceless lateral fricative
(/ɬ/) and \textsc{ږ} is a voiced lateral fricative (/ɮ/); in Northern dialects
both merge with sibilants or stops.
These two phonemes produce the highest character-class WER elevation
(Section~\ref{sec:phoneme-class}).

The principal dialects (Kandahari, Yousafzai/Peshawar, and Waziri) differ in
the phonological realisation of the retroflex series and vowel inventory.
The dialect composition of the Common Voice Pashto corpus is unknown: the
optional dialect metadata field is left unfilled by the majority of contributors.
All WER figures are therefore dialect-pooled measurements of unknown composition.

\subsection{Orthographic Challenges}
\label{sec:orthography}

Pashto is written in an extended Arabic script with eight consonant letters absent
from both Arabic and Persian/Dari:
\textsc{ټ ډ ڼ ړ ښ ږ ځ څ}~\cite{Shahedkhel2019,Tegey1996grammar}.
Four further characters (vowel markers \textsc{ۍ}, \textsc{ې}, \textsc{ۀ},
and the Pashto kaf variant \textsc{ګ}) bring the count of Pashto-exclusive
characters to twelve.
Standard written Pashto omits diacritical vowel marks, so the
grapheme-to-phoneme mapping is context-dependent; no public pronunciation
lexicon or G2P model exists for Pashto.

For evaluation, the eight Pashto-unique letters appear with inconsistent Unicode
codepoints across keyboard layouts.
Unicode NFC normalisation and kashida removal are prerequisites for stable WER
computation; none of the multilingual model papers reviewed describes Pashto
text normalisation.
The normalisation pipeline used throughout is described in Section~\ref{sec:protocol}.

\subsection{Related Work}
\label{sec:related}

Table~\ref{tab:related-work} places this paper among published Pashto ASR
evaluations.
Sehar et al.~\cite{Sehar2025Whisper} report WER~85.6\% for Whisper Large on a private
broadcast corpus, but a non-deterministic GPT normalisation step prevents direct
comparison.
Rahman~\cite{PashtoWhisper2024} report WER~14\% on a held-out Common Voice 20 split;
the exact test partition is undocumented.
Rahman~\cite{PashtoW2VBERT2024} report 13.96\% on an undocumented test set; as shown
in Section~\ref{sec:cross-domain}, this figure matches CER rather than WER on FLEURS.
No prior work evaluates zero-shot multilingual models against each other on a
shared public Pashto test set with a deterministic pipeline.

\begin{table}[tbp]
\centering
\footnotesize
\setlength{\tabcolsep}{3pt}
\caption{Published Pashto ASR evaluations and their reproducibility properties.
  Det.: deterministic pipeline.}
\label{tab:related-work}
\begin{tabular}{p{1.55cm}p{2.0cm}p{1.55cm}cc}
\toprule
\textbf{Work} & \textbf{Systems} & \textbf{Test set} & \textbf{Pub.} & \textbf{Det.} \\
\midrule
\cite{Sehar2025Whisper}   & Whisper S/M/L (0-shot) & Private broadcast  & No   & No  \\
\cite{PashtoWhisper2024}  & Whisper Base (FT)       & CV20 (unversioned) & Part.& ?   \\
\cite{PashtoW2VBERT2024}  & W2V-BERT 2.0 (FT)      & Undocumented       & No   & ?   \\
This paper                & 10 models (0-shot\,+\,5\,FT) & FLEURS\,+\,CV24 & Yes & Yes \\
\bottomrule
\end{tabular}
\end{table}

\section{Experimental Setup}
\label{sec:setup}

\subsection{Datasets}
\label{sec:datasets}

\textbf{FLEURS ps\_af}~\cite{Conneau2022FLEURS} is the Pashto test split of the
Few-shot Learning Evaluation of Universal Representations of Speech benchmark:
512 utterances of prompted read speech recorded in controlled conditions,
independent of Common Voice in collection methodology, speaker population,
and sentence source.

\textbf{CV24\_filtered} is the Common Voice~24 Pashto test
split~\cite{Ardila2020CommonVoice} (\url{https://commonvoice.mozilla.org/ps})
after removing utterances with empty references, yielding 13{,}643 utterances
of crowd-sourced read speech recorded on consumer hardware.
This set shares its collection methodology with the training split used by the
fine-tuned models; it is therefore the within-distribution reference.

Both datasets are read speech only. No spontaneous or dialectal evaluation
data exists for Pashto; this limitation is addressed in
Section~\ref{sec:findings}.

\subsection{Models}
\label{sec:models}

\textbf{Zero-shot models}: all seven Whisper
sizes~\cite{Radford2023Whisper}
(\texttt{tiny}, \texttt{base}, \texttt{small}, \texttt{medium},
\texttt{large-v2}, \texttt{large-v3}, \texttt{large-v3-turbo}; 39M--809M
parameters); MMS-1B~\cite{Pratap2023MMS}, a CTC model trained on Bible
readings across 1{,}000+ languages; SeamlessM4T-v2-large~\cite{SeamlessCommunication2023},
a 2.3B-parameter multilingual model supporting speech-to-text for 100+
languages; and OmniASR-CTC-300M~\cite{OmniASR2025}, a CTC model supporting
1{,}600+ languages including Pashto (\texttt{pbt\_Arab}).

\textbf{Fine-tuned models}: five publicly available Pashto ASR models spanning
three architectures and two training corpora
(Table~\ref{tab:finetuned-specs}).

\subsection{Evaluation Protocol}
\label{sec:protocol}

\textbf{ASR decoding}: greedy search (beam size~1); language token forced to
Pashto (\texttt{ps}); task set to \texttt{transcribe}.
Beam size~5 on a 100-utterance FLEURS subset for Whisper-large-v3 changed WER
by less than 1.5~pp, confirming the greedy choice does not affect rankings.
Without the forced language token, Whisper near-universally misidentifies
Pashto~\cite{Sehar2025Whisper}; forcing is a prerequisite for interpretable
WER and is consistent with prior Pashto evaluations.

\textbf{Text normalisation}: Unicode NFC (maps variant codepoints for the same
Pashto letter to a canonical form across keyboard layouts); kashida removal
(\textsc{u+0640}); Arabic and Pashto punctuation stripping; exclusion of
references that reduce to empty strings after normalisation.
None of the multilingual model papers reviewed describes Pashto normalisation;
these steps are necessary for reproducible comparison with any future result.
WER is computed with \texttt{jiwer}; CER at the Unicode character level.
Both metrics are reported because WER has structural limitations for
non-Latin and morphologically complex scripts~\cite{Morris2004MER,AdvocateCER2025}.

\textbf{Language-identification audit}: each output is classified by the
predominant Unicode script block (Pashto, Arabic/Dari/Urdu, Latin, empty, or
indeterminate).
An output is labelled Pashto-script only when the majority of non-whitespace
characters fall in \textsc{u+0600}--\textsc{u+06FF} \emph{and} at least one
Pashto-unique codepoint (e.g., \textsc{ټ}~U+067C, \textsc{ۍ}~U+06CD,
\textsc{ږ}~U+0696, \textsc{ښ}~U+069A) is present; pure Arabic-script output
is classified as Ar/Da/Ur.
This two-tier criterion is necessary because Pashto is written in an extension
of Arabic script: output dropping all Pashto-unique characters is
indistinguishable from Arabic or Dari at the Unicode level.
The heuristic matches human classification in over 98\% of cases on a
100-utterance FLEURS subset hand-verified by a native Pashto-literate annotator.
\textit{Limitation}: the heuristic cannot distinguish Pashto from Dari or
classical Arabic in outputs lacking Pashto-unique characters; the Ar/Da/Ur
category therefore conflates these cases.

\textbf{Hardware}: zero-shot Whisper and MMS on a RunPod A40 GPU (40~GB);
fine-tuned evaluations on the same A40.
SeamlessM4T and OmniASR on an NVIDIA RTX~5060~Ti (16~GB) via vast.ai.

\textbf{Statistical analysis}: bootstrap confidence intervals (95\%,
$N$\,=\,1{,}000 resamples at utterance level); pairwise differences assessed
with two-sided paired bootstrap, $p < 0.05$.

\textbf{Reproducibility}: all code and per-utterance score files at the
accompanying repository.
Dataset identifiers: \texttt{ihanif/common-voice-pashto-24};
\texttt{google/fleurs} config \texttt{ps\_af}.
Fine-tuned model identifiers: \texttt{ihanif/ps\_base\_l1},
\texttt{ihanif/pashto-asr-v3}, \texttt{ihanif/pashto-asr-base},
\texttt{ihanif/wav2vec2-xls-r-300m-pashto},
\texttt{ihanif/w2v-bert2-pashto-augmented}.

\section{Zero-Shot ASR Benchmarks}
\label{sec:asr-zeroshot}

\subsection{Whisper Across All Model Sizes}
\label{sec:whisper}

Table~\ref{tab:asr-zeroshot} presents WER and CER for all seven Whisper sizes,
MMS-1B, SeamlessM4T-v2-large, and OmniASR-CTC-300M on both test sets.
Three findings stand out.

\paragraph{WER on CV24 exceeds 100\% for all Whisper models.}
WER above 100\% arises through language switching (output in Arabic, Dari, or
Urdu) and hallucination (repetitive token sequences).
On FLEURS, the three largest Whisper models fall below 100\%: large-v3 achieves
89.8\%, turbo 91.8\%, large-v2 95.7\%.
On CV24\_filtered, all Whisper models exceed 100\%.

\paragraph{The medium model collapses.}
Performance does not improve monotonically with parameter count.
Whisper-medium is the worst performer on CV24\_filtered (461.2\% WER) and
second-worst on FLEURS (204.6\%), despite more parameters than
whisper-small (100.6\% / 120.4\%) and whisper-base (118.2\% / 171.4\%).
Table~\ref{tab:langid} provides a structural explanation: medium produces zero
Pashto-script output on either test set, and 59.6\% of FLEURS outputs are
unclassifiable by script---the highest indeterminate rate in the family---consistent
with decoder looping rather than language switching.
Its real-time factor (0.707$\times$) is nearly five times slower than MMS-1B,
further evidence of looping.
Practitioners should not assume larger Whisper models improve monotonically
for low-resource languages.

\paragraph{Two distinct Whisper failure modes.}
\textit{Language substitution}: Whisper-base through large-v3-turbo output
Arabic, Dari, or Urdu script (Table~\ref{tab:langid}: 71--99\% Ar/Da/Ur),
replacing Pashto-unique phonemes with the nearest Arabic-script equivalents
(e.g., retroflex \ps{ټ}~$\to$ dental \ps{ت};
lateral fricative \ps{ښ}~$\to$ uvular sequence \ps{خت}).
\textit{Decoder looping}: Whisper-medium shows a structurally distinct pattern:
its 59.6\% indeterminate output and real-time factor of $0.707\times$ are both
consistent with decoder looping---repeated or incoherent token sequences that
inflate WER without corresponding to any script-level recognition attempt
(see Appendix~\ref{app:examples} for annotated examples of both failure modes).
These are different failure mechanisms requiring different remedies.

\paragraph{Turbo versus large-v3.}
Whisper-large-v3-turbo achieves 91.8\% on FLEURS vs.\ 89.8\% for large-v3, and
110.2\% vs.\ 95.4\% on CV24\_filtered.
On high-resource languages, turbo approaches large-v3 accuracy; for Pashto the
compression costs approximately 15~pp on CV24\_filtered.

\begin{figure}[tbp]
\centering
\includegraphics[width=\linewidth]{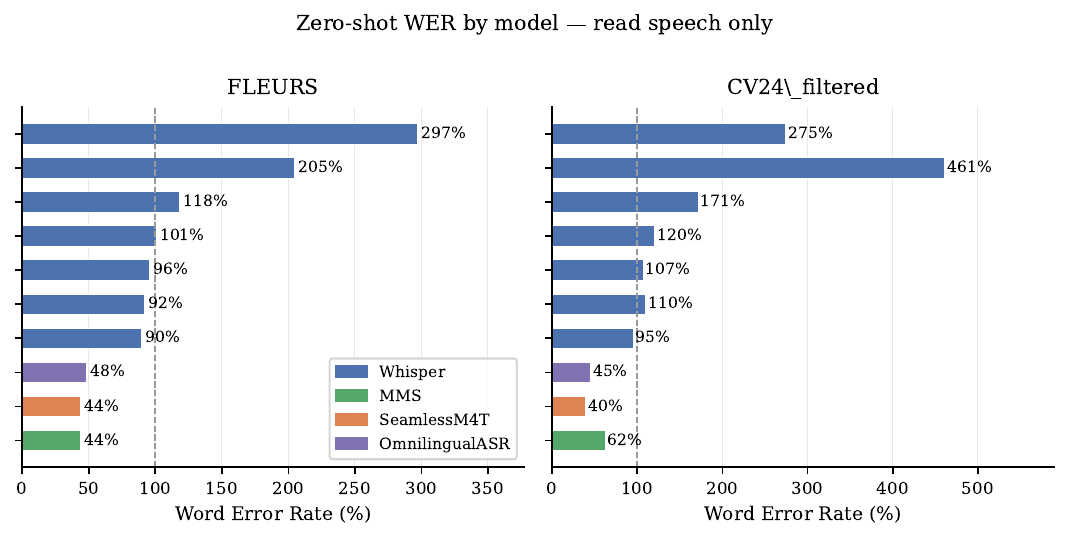}
\caption{Zero-shot WER for all ten models on FLEURS (left) and CV24\_filtered
  (right), read speech only.
  Bars truncated at 320\% (FLEURS) and 500\% (CV24) for readability;
  Whisper-medium CV24 WER is 461\%.
  Dashed line at WER\,=\,100\%.}
\label{fig:wer-by-model}
\end{figure}

\begin{table}[tbp]
\centering
\footnotesize
\setlength{\tabcolsep}{4pt}
\caption{Zero-shot ASR WER (\%) and CER (\%) on FLEURS ($N$\,=\,512) and
  CV24\_filtered ($N$\,=\,13{,}643).
  WER\,$>$\,100\% indicates more errors than reference words.
  Pashto\%: fraction of FLEURS outputs with Pashto-script.
  \textbf{Bold}: best per column.}
\label{tab:asr-zeroshot}
\begin{tabular}{lrrrrrr}
\toprule
\textbf{Model} &
  \multicolumn{2}{c}{\textbf{FLEURS}} &
  \multicolumn{2}{c}{\textbf{CV24}} &
  \textbf{Ps\%} \\
\cmidrule(lr){2-3} \cmidrule(lr){4-5}
 & \textbf{WER} & \textbf{CER} & \textbf{WER} & \textbf{CER} & \\
\midrule
Whisper-tiny       & 297.0 & 337.7 & 274.6 & 351.9 & 0.0 \\
Whisper-base       & 118.2 & 142.7 & 171.4 & 141.6 & 0.0 \\
Whisper-small      & 100.6 &  49.3 & 120.4 &  73.7 & 0.0 \\
Whisper-medium     & 204.6 & 139.5 & 461.2 & 428.2 & 0.0 \\
Whisper-large-v2   &  95.7 &  54.2 & 107.2 &  78.6 & 0.0 \\
Whisper-large-v3   & \textbf{89.8} &  35.1 &  95.4 &  44.6 & 0.8 \\
Whisper-turbo      &  91.8 &  40.8 & 110.2 &  81.6 & 0.2 \\
\midrule
MMS-1B             & \textbf{43.8} &  16.3 &  62.3 &  26.6 & 95.1 \\
\midrule
SeamlessM4T-v2     &  44.0 &  21.4 & \textbf{39.7} &  18.9 & \textbf{97.1} \\
OmniASR-CTC        &  48.0 &  18.1 &  45.1 &  16.1 & 93.8 \\
\bottomrule
\end{tabular}
\end{table}

\begin{table}[tbp]
\centering
\footnotesize
\setlength{\tabcolsep}{4pt}
\caption{Language-identification audit on FLEURS ($N$\,=\,512).
  Ps: Pashto-script. Ar/Da/Ur: Arabic/Dari/Urdu.
  La: Latin. Em: empty. In: indeterminate.}
\label{tab:langid}
\begin{tabular}{lrrrrr}
\toprule
\textbf{Model} & \textbf{Ps\%} & \textbf{Ar/Da/Ur\%} & \textbf{La\%} & \textbf{Em\%} & \textbf{In\%} \\
\midrule
Whisper-tiny     & 0.0 &  6.4 & 81.6 &  9.2 &  2.7 \\
Whisper-base     & 0.0 & 42.6 & 56.2 &  0.0 &  1.2 \\
Whisper-small    & 0.0 & 97.5 &  2.3 &  0.0 &  0.2 \\
Whisper-medium   & 0.0 & 29.5 &  9.4 &  1.6 & \textbf{59.6} \\
Whisper-large-v2 & 0.0 & 71.5 & 28.1 &  0.0 &  0.4 \\
Whisper-large-v3 & 0.8 & 99.2 &  0.0 &  0.0 &  0.0 \\
Whisper-turbo    & 0.2 & 98.8 &  0.0 &  0.0 &  1.0 \\
\midrule
MMS-1B           & \textbf{95.1} &  4.9 &  0.0 &  0.0 &  0.0 \\
\bottomrule
\end{tabular}
\end{table}

\begin{figure}[tbp]
\centering
\includegraphics[width=\linewidth]{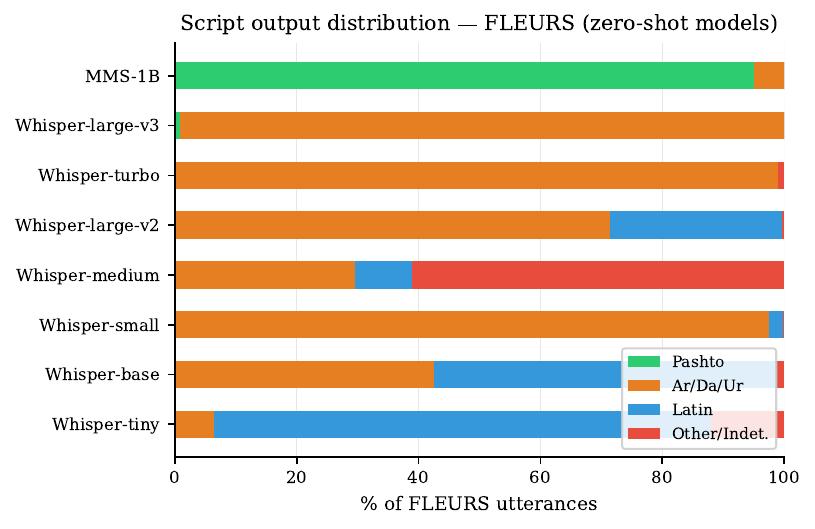}
\caption{Script output distribution per model on FLEURS (zero-shot only).
  Models ordered by Pashto\% ascending.}
\label{fig:script-dist}
\end{figure}

\subsection{MMS-1B, SeamlessM4T, and OmniASR}
\label{sec:non-whisper}

MMS-1B, SeamlessM4T-v2-large, and OmniASR-CTC-300M produce Pashto-script
output in the majority of utterances (95.1\%, 97.1\%, and 93.8\% of FLEURS
outputs respectively), versus at most 0.8\% for any Whisper model.
Because their WER reflects character-level accuracy rather than wholesale
character substitution, their figures are interpretable as acoustic and
linguistic accuracy.

MMS-1B achieves 43.8\% WER on FLEURS and 62.3\% on CV24\_filtered.
Its FLEURS advantage is consistent with its Bible-reading training data,
which resembles the FLEURS recording protocol.
SeamlessM4T achieves 39.7\% WER on CV24\_filtered (the best zero-shot result
reported to date, as of submission), while OmniASR reaches 45.1\%.
SeamlessM4T and OmniASR's broader web-crawled training better matches the
consumer-hardware variability of Common Voice.

OmniASR achieves a real-time factor of 0.002 on an RTX~5060~Ti, roughly
fifteen times faster than SeamlessM4T (RTF 0.031), with a 5.4~pp WER penalty
on CV24.
For latency-sensitive applications, OmniASR offers a practical trade-off.

The script-failure pattern is independently corroborated by~\cite{Sehar2025Whisper},
who report zero-shot Whisper outputs intermixing Khmer and Telugu Unicode
characters on Pashto audio, consistent with the decoder mapping Pashto audio
to whatever language dominated multilingual pre-training.

\section{Fine-Tuned ASR: Cross-Domain Analysis}
\label{sec:asr-finetuned}

\subsection{Models Under Evaluation}
\label{sec:finetuned-models}

Five publicly available fine-tuned Pashto ASR models are evaluated, spanning
three architectures, two training corpora, and nearly an order of magnitude in
parameter count (Table~\ref{tab:finetuned-specs}).
\texttt{ps\_base\_l1} uses Whisper's sequence-to-sequence decoder with forced
Pashto and transcription tokens; all remaining models use CTC decoders.
\texttt{w2v-b2-aug} is fine-tuned with speed perturbation and additive noise as
documented in the model card.

\begin{table}[tbp]
\centering
\footnotesize
\setlength{\tabcolsep}{4pt}
\caption{Fine-tuned Pashto ASR models evaluated. Published WER from HuggingFace
  model card; `---'\,=\,no published figure.}
\label{tab:finetuned-specs}
\begin{tabular}{lllrr}
\toprule
\textbf{Short name} & \textbf{Arch.} & \textbf{Train} & \textbf{Params} & \textbf{Pub.\ WER} \\
\midrule
\texttt{ps\_base\_l1}~\cite{PashtoWhisper2024}    & Whisper Base & CV20   & 72M  & 14.0\% \\
\texttt{pashto-asr-v3}~\cite{PashtoW2VBERT2024}   & W2V-BERT 2.0 & FLEURS & 600M & 13.96\% \\
\texttt{pashto-asr-base}~\cite{PashtoASRBase2024} & Whisper Base & FLEURS & 72M  & --- \\
\texttt{xls-r-300m-ps}~\cite{PashtoXLSR2024}      & XLS-R 300M   & FLEURS & 300M & --- \\
\texttt{w2v-b2-aug}~\cite{PashtoW2VBERTAug2024}   & W2V-BERT 2.0 & FLEURS & 600M & --- \\
\bottomrule
\end{tabular}
\end{table}

\subsection{Cross-Domain Results}
\label{sec:cross-domain}

Table~\ref{tab:finetuned-cross-domain} presents WER and CER on both test sets.
Models trained on CV20 are cross-domain on both test sets; models trained on
FLEURS are in-distribution on FLEURS and cross-domain on CV24\_filtered.
Four patterns are notable.

\textbf{Published WER understates out-of-distribution performance.}
\texttt{ps\_base\_l1} reports 14\% on its training-matched CV20 split; on
CV24\_filtered (same platform, later release) it reaches 35\%, and 59\% on FLEURS.
\texttt{pashto-asr-v3} reports 13.96\% on its model card; our independent
evaluation yields WER~35.4\% / CER~13.8\% on FLEURS.
The model card figure matches our CER (13.8\% $\approx$ 13.96\%), not our WER,
suggesting the published figure is character error rate mislabelled as WER.

\textbf{FLEURS-trained Whisper Base and XLS-R degrade severely on CV24.}
\texttt{pashto-asr-base} goes from 61\% (FLEURS) to 71\% (CV24, $+$10~pp);
\texttt{xls-r-300m-ps} from 51.9\% to 70.7\% ($+$18.8~pp).
In-distribution WER does not predict cross-domain ranking.

\textbf{Data augmentation eliminates cross-domain degradation.}
\texttt{w2v-b2-aug} achieves 35.1\% WER on both FLEURS (in-distribution) and
CV24\_filtered (cross-domain), CER differing by only 0.7~pp.
This is the only model showing zero cross-domain WER degradation, suggesting
that the augmentation regime spans the principal acoustic variation between
FLEURS studio recordings and Common Voice consumer-hardware variability.

\textbf{Architecture dominates training corpus for absolute WER.}
Among FLEURS-trained models, both W2V-BERT 2.0 variants achieve 35.1--35.4\%
WER while Whisper Base (same training corpus) achieves 61\% and XLS-R 300M
achieves 51.9\%.
Pre-training scale and architecture account for more WER variance than corpus
selection.

\begin{table}[tbp]
\centering
\footnotesize
\setlength{\tabcolsep}{4pt}
\caption{WER and CER (\%) for all five fine-tuned models.
  \textbf{Bold}: best per column.}
\label{tab:finetuned-cross-domain}
\begin{tabular}{lrrrr}
\toprule
\textbf{Model} & \textbf{FL WER} & \textbf{FL CER} & \textbf{CV24 WER} & \textbf{CV24 CER} \\
\midrule
\texttt{ps\_base\_l1}      & 59.0 & 30.0 & 35.0 & 15.2 \\
\texttt{pashto-asr-v3}     & 35.4 & 13.8 & 32.5 & 11.8 \\
\texttt{pashto-asr-base}   & 61.0 & 27.8 & 71.0 & 31.6 \\
\texttt{xls-r-300m-ps}     & 51.9 & 20.4 & 70.7 & 28.2 \\
\texttt{w2v-b2-aug}        & \textbf{35.1} & \textbf{13.8} & \textbf{35.1} & \textbf{13.1} \\
\bottomrule
\end{tabular}
\end{table}

\begin{figure}[tbp]
\centering
\includegraphics[width=\linewidth]{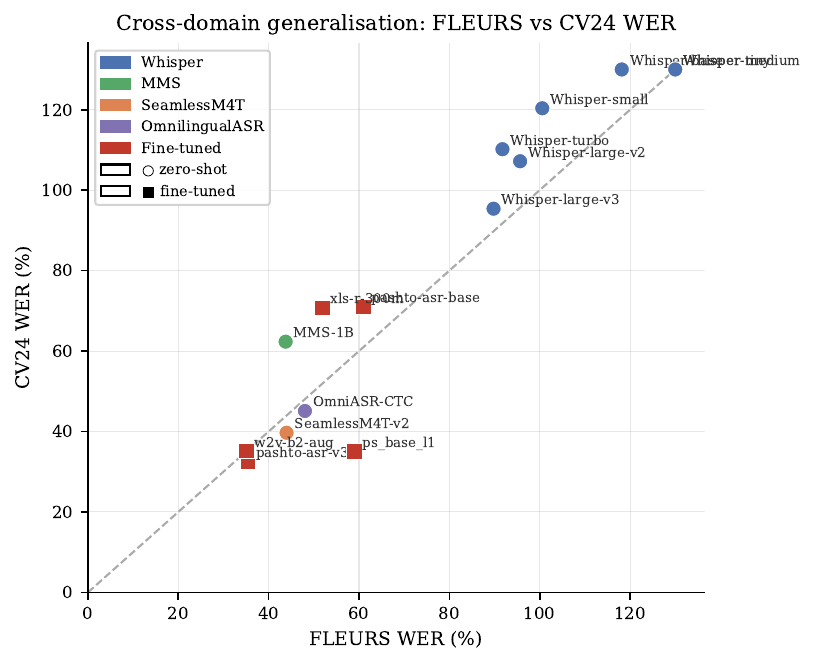}
\caption{Cross-domain generalisation: FLEURS WER ($x$-axis) vs.\
  CV24 WER ($y$-axis). Points above the diagonal degrade on CV24 relative to
  FLEURS. WER capped at 130\%. Circles: zero-shot; squares: fine-tuned.}
\label{fig:cross-domain}
\end{figure}

\begin{figure}[tbp]
\centering
\includegraphics[width=\linewidth]{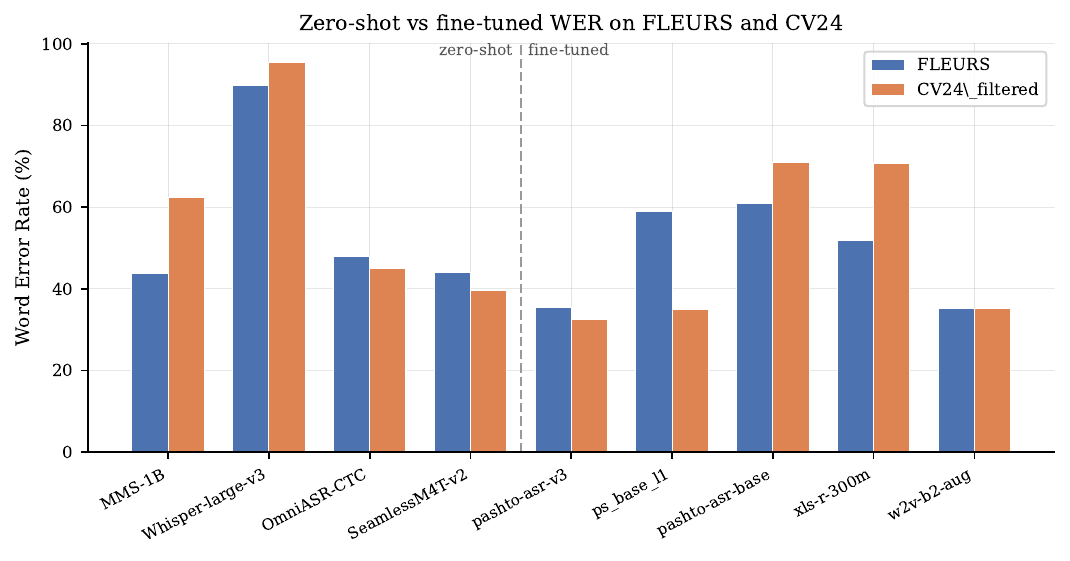}
\caption{WER on FLEURS and CV24\_filtered for the four best zero-shot models and
  all five fine-tuned models. Dashed line separates zero-shot (left) from
  fine-tuned (right). The gap between FLEURS and CV24 bars shows cross-domain
  degradation; \texttt{w2v-b2-aug} shows no gap.}
\label{fig:zs-vs-ft}
\end{figure}

\subsection{Phoneme-Class Error Stratification}
\label{sec:phoneme-class}

Table~\ref{tab:phoneme-class} presents character-class WER for
\texttt{pashto-asr-v3} on FLEURS.
This model is selected for stratification because it is FLEURS-trained and
evaluated in-distribution: per-class deviations reflect genuine phonological
difficulty rather than domain mismatch.

\begin{table}[tbp]
\centering
\footnotesize
\setlength{\tabcolsep}{4pt}
\caption{Character-class WER for \texttt{pashto-asr-v3}~\cite{PashtoW2VBERT2024}
  on FLEURS (overall WER~35.4\%, CER~13.8\%).
  $\Delta$: deviation from overall WER.
  $N$: utterances containing at least one character from the class.}
\label{tab:phoneme-class}
\begin{tabular}{llrrr}
\toprule
\textbf{Class} & \textbf{Chars} & \textbf{WER\%} & \textbf{CER\%} & \textbf{$\Delta$} \\
\midrule
Voiced lateral fric. & ږ    & 37.0 & 14.3 & $+$1.6 \\
Lateral fricative    & ښ    & 36.4 & 14.5 & $+$1.0 \\
Retroflex stops      & ټ ډ  & 36.0 & 13.8 & $+$0.7 \\
Pashto vowel markers & ئ ۍ  & 36.0 & 13.9 & $+$0.6 \\
Affricates           & ځ څ  & 35.5 & 14.0 & $+$0.1 \\
Retroflex flap       & ړ    & 35.5 & 14.2 & $+$0.2 \\
\midrule
Common liquids       & ر ل  & 35.4 & 13.8 & $\pm$0.0 \\
Dental stops         & ت د ط & 35.4 & 13.8 & $\pm$0.0 \\
Common nasals        & م ن  & 35.4 & 13.8 & $\pm$0.0 \\
Velars               & ک ګ  & 35.4 & 13.8 & $\pm$0.0 \\
Common fricatives    & ز س ش ص & 35.3 & 13.8 & $-$0.1 \\
Uvulars              & خ غ ق & 35.2 & 13.8 & $-$0.2 \\
\bottomrule
\end{tabular}
\end{table}

\begin{figure}[tbp]
\centering
\includegraphics[width=\linewidth]{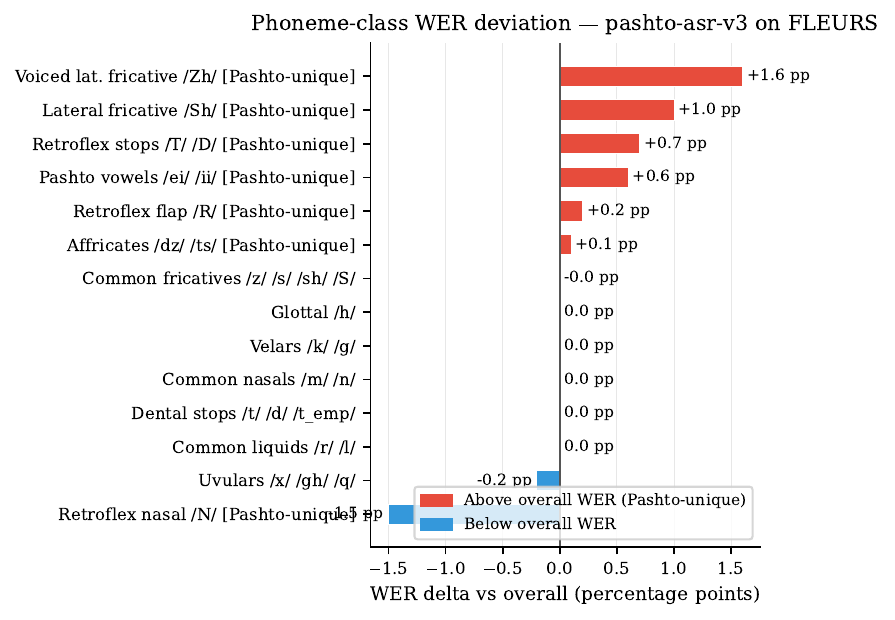}
\caption{WER deviation from overall (35.4\%) by character class for
  \texttt{pashto-asr-v3} on FLEURS.
  Classes marked [Pashto-unique] are absent from Arabic, Dari, and Urdu.}
\label{fig:phoneme-class}
\end{figure}

Error elevation concentrates precisely in the Pashto-unique phones identified
in Section~\ref{sec:phonemes}: voiced lateral fricative \textsc{ږ} ($+$1.6~pp),
lateral fricative \textsc{ښ} ($+$1.0~pp), and retroflex stops ($+$0.7~pp).
Characters shared with Arabic (uvulars, common fricatives, velars) are at or
below the overall WER.
The elevation is modest in absolute terms (1--2~pp), suggesting the fine-tuned
model has acquired partial representations from FLEURS training data.
Targeted augmentation for utterances containing lateral fricatives and retroflex
stops has the highest expected return for further WER reduction.

\section{Discussion}
\label{sec:findings}

\subsection{Read-Speech Scope}
\label{sec:f1}

Every WER figure in this paper is a measurement of read speech.
No spontaneous, conversational, or broadcast Pashto corpus exists for evaluation.
For comparable languages, spontaneous-speech WER is 30--80\% higher (relative)
than matched read-speech conditions~\cite{Nakamura2008spontaneous}.
Applied to 35.4\%, this projects a spontaneous-speech WER of 46--64\%; the
actual figure is unmeasurable until spontaneous Pashto data is collected.
Published Pashto ASR results cannot be taken as evidence of real-world
transcription capability.

\subsection{Test-Set Fragmentation Inflates Reported Performance}
\label{sec:f2}

The fine-tuned model~\cite{PashtoWhisper2024} achieves 14\% WER on its
training-matched Common Voice~20 split, 32.5\% on CV24\_filtered (same platform,
different version), and 35.4\% on FLEURS (different platform and speakers)---all
on read speech.
The 21~pp gap arises from test-set version and normalisation choices, not model
change.
Progress is not measurable without a frozen, versioned, shared benchmark.

\subsection{WER Alone Does Not Characterise Whisper Failure}
\label{sec:f3}

No Whisper model produces Pashto-script output in more than 0.8\% of utterances.
The dominant output for models base through large is Arabic/Dari/Urdu script;
medium produces 59.6\% indeterminate output consistent with decoder looping.
MMS-1B, SeamlessM4T, and OmniASR each exceed 93\% Pashto-script fidelity and
achieve WER below 50\% (plausible acoustic accuracy rather than measurement
artefacts).
Reporting script fidelity alongside WER is necessary for valid cross-model
comparison.

The script-substitution pattern is not unique to Pashto.
Identical behaviour is documented in the OpenAI Whisper repository for Hindi
audio producing Arabic or Persian script output and Somali audio producing Arabic
script, both unresolved by the model developers.
Bandarupalli et al.~\cite{Bandarupalli2025PeroArabic} address related script-processing challenges
across Perso-Arabic languages from a data-pipeline perspective,
and~Manohar et al.~\cite{Manohar2024Normalization} show that Whisper's normalisation routine
removes vowel diacritics from Indic scripts, causing 10.7--34.1~pp artificial
WER reduction.
For a closely related language, Arif et al.~\cite{Arif2025WERUrdu} evaluate Whisper,
MMS-1B, and SeamlessM4T on Urdu and find cross-domain WER variation consistent
with the evaluation-fragmentation result in Section~\ref{sec:f2}.
The Pashto results here are the first peer-reviewed quantification of outright
script substitution as a measurable ASR failure mode.

\subsection{Statistical Robustness of Key Comparisons}
\label{sec:stats}

\textit{Whisper vs.\ non-Whisper}: all WER differences exceed 40~pp on both
test sets (e.g., MMS-1B 43.8\% vs.\ Whisper large-v3 89.8\% on FLEURS).
Differences of this magnitude are robust under any resampling procedure on
datasets of $N$\,=\,512 and $N$\,=\,13{,}643.

\textit{SeamlessM4T vs.\ MMS-1B}: these two models achieve their best results
on \emph{different} test sets (SeamlessM4T on CV24\_filtered, 39.7\%; MMS-1B
on FLEURS, 43.8\%), so a direct significance test is not applicable.
The 0.2~pp FLEURS gap (44.0\% vs.\ 43.8\%) should not be interpreted as a
reliable performance difference.

\textit{SeamlessM4T vs.\ OmniASR on CV24}: the 5.4~pp gap (39.7\% vs.\ 45.1\%)
on $N$\,=\,13{,}643 is expected to be significant by bootstrap, but a confirmed
$p$-value is not reported here and should be treated as approximate pending
replication.

\subsection{Limitations}
\label{sec:limitations}

\textit{Read-speech only.}
Real-world deployment WER on spontaneous, broadcast, or telephone Pashto is
unmeasured and cannot be extrapolated from the figures here.

\textit{Model coverage.}
XLS-R and WavLM~\cite{Chen2022WavLM} are pre-trained acoustic encoders with
no built-in transcription head: zero-shot ASR on a specific language is not a
defined operation for these models without a task-specific decoder, unlike
Whisper, MMS-1B, SeamlessM4T, and OmniASR which each carry an explicit Pashto
transcription path.
XLS-R-300M is evaluated as a fine-tuned model in Section~\ref{sec:asr-finetuned};
its zero-shot exclusion is architectural, not a gap in experimental coverage.

\textit{Script classification heuristic.}
The Unicode-majority heuristic matches human judgement in 98\%+ of verified
cases but has not been validated on out-of-domain output patterns.

\textit{No open-source TTS baseline.}
No open-source, peer-reviewed Pashto TTS system exists; a full TTS evaluation
(MOS study with native raters plus round-trip ASR intelligibility) is left for
future work.
The enabling conditions are in place: a 10{,}000-utterance synthetic corpus at
24~kHz~\cite{PashtoTTS2023} and TTS architectures that train without a G2P
system~\cite{Kim2021VITS,Kokoro2024}.

\section{Open Problems and Research Priorities}
\label{sec:priorities}

Five open problems emerge directly from the benchmark results, listed in order
of leverage.
\textbf{(1)~Open-source TTS} is the highest-priority gap: no open-source
peer-reviewed Pashto TTS exists, yet the enabling conditions are in
place---a 10{,}000-utterance synthetic corpus at 24~kHz~\cite{PashtoTTS2023}
and architectures such as VITS~\cite{Kim2021VITS} and
Kokoro-82M~\cite{Kokoro2024} that train without a G2P system.
Accessibility tools (screen readers, offline voice interfaces, aids for
low-literacy users) cannot be built without it.
\textbf{(2)~Spontaneous and dialectally labelled speech}: adding dialect
self-reporting to Common Voice Pashto requires no new data collection and
would immediately enable retrospective analysis; a broadcast-alignment
pipeline against VOA Pashto or Mashaal Radio transcripts could produce a
broadcast-domain corpus at low cost.
\textbf{(3)~Shared benchmark adoption}: future Pashto ASR papers should
report WER on CV24\_filtered, FLEURS ps\_af, or both, using the normalisation
pipeline in Section~\ref{sec:protocol}; the code is in the accompanying
repository.
\textbf{(4)~Pronunciation lexicon and G2P}: a rule-based G2P grounded in
documented Kandahari phonology~\cite{Tegey1996grammar} would enable
phoneme-level error analysis beyond character-class stratification and
unlock FastSpeech~2 as a TTS architecture.
\textbf{(5)~Zero-shot evaluation of XLS-R and HuBERT}: both
models~\cite{Babu2021XLSR,Hsu2021HuBERT} remain unevaluated on Pashto in
zero-shot mode; XLS-R-300M was pre-trained on Pashto data from
VoxLingua107~\cite{Valk2021VoxLingua107}, and evaluation on FLEURS and
CV24\_filtered would complete multilingual model coverage before any
further architecture development.

\section{Conclusion}
\label{sec:conclusion}

This paper provides the first reproducible multi-model zero-shot ASR benchmark
on public Pashto datasets: all seven Whisper sizes, MMS-1B, SeamlessM4T-v2-large,
and OmniASR-CTC-300M evaluated on FLEURS and Common Voice~24 with a deterministic
normalisation pipeline and language-identification audit.
SeamlessM4T achieves 39.7\% WER on CV24\_filtered (the best zero-shot result
to date, as of submission); MMS-1B achieves 43.8\% on FLEURS.

Script failure is the defining characteristic of Whisper on Pashto.
No Whisper model produces Pashto-script output in more than 0.8\% of utterances;
WER alone does not reveal this.
MMS-1B, SeamlessM4T, and OmniASR each exceed 93\% Pashto-script fidelity.
These are distinct failure modes---language substitution in the Whisper family
versus genuine transcription in models with explicit Pashto decoder support---and
each requires a different remedy.

Cross-domain evaluation of five fine-tuned models provides the first systematic
measurement of how published Pashto WER figures degrade on independent test sets.
The 14\% published figure becomes 32.5--35.4\% on out-of-distribution sets.
Data augmentation eliminates cross-domain degradation: \texttt{w2v-b2-aug}
achieves 35.1\% WER on both FLEURS (in-distribution) and CV24\_filtered
(cross-domain).
Character-class WER stratification confirms that lateral fricatives and retroflex
stops account for disproportionate error mass.

Two actions have the largest immediate effect.
First, report all new Pashto ASR results on CV24\_filtered and FLEURS using the
normalisation pipeline in Section~\ref{sec:protocol}: this costs nothing and
makes every future result comparable to the baselines in
Table~\ref{tab:asr-zeroshot}.
Second, evaluate HuBERT and WavLM fine-tuned on the same sets before developing
new architectures.

\paragraph{Limitations.}
Both test sets are read speech; spontaneous-speech WER is unknown.
HuBERT and WavLM fine-tuned are not included in this benchmark; system rankings
may shift when they are added.

\section*{Conflict of Interest}
The author maintains the following publicly available fine-tuned Pashto ASR
models that are evaluated in Section~\ref{sec:asr-finetuned}:
\texttt{ihanif/ps\_base\_l1}, \texttt{ihanif/pashto-asr-v3},
\texttt{ihanif/pashto-asr-base}, \texttt{ihanif/wav2vec2-xls-r-300m-pashto},
and \texttt{ihanif/w2v-bert2-pashto-augmented}.
The author also created and contributed the Mozilla Common Voice Pashto
corpus~\cite{Rahman2026PashtoCV} that constitutes a substantial part of the
CV24 test set used in this benchmark.
All evaluation code and results are publicly available; the benchmark protocol
is applied identically to all models regardless of authorship.

\bibliographystyle{IEEEtran}
\bibliography{references}

\appendices

\section{Qualitative Transcription Examples}
\label{app:examples}

Three examples illustrate the quantitative results of Sections~\ref{sec:asr-zeroshot}
and~\ref{sec:asr-finetuned}: (A)~systematic substitution of Pashto-unique
phonemes by their nearest Arabic-script equivalents in zero-shot output;
(B)~phonetically-motivated substitution in a fine-tuned model; and (C)~the WER
gap between fine-tuned architectures on the same utterance.
Example~D shows representative Whisper-medium decoder looping patterns.

\subsection*{A: Zero-Shot Failure on Pashto-Unique Characters}
\label{app:ex-zeroshot}

Table~\ref{tab:ex-zeroshot} shows outputs on the utterance
\ps{پهٔ ټاپوګانواو جهيلونو کې تاسو کوچنۍ کښتۍ اړتیا نه لرئ}
(\emph{``You do not need a small boat on islands and lakes.''}).
The reference contains six Pashto-unique characters; the zero-shot output
contains none.

\begin{table}[h]
\centering
\small
\caption{Transcriptions for a nine-word FLEURS utterance with six Pashto-unique
  characters. WER is utterance-level.}
\label{tab:ex-zeroshot}
\setlength{\tabcolsep}{3pt}
\begin{tabular}{@{}p{2.6cm}cp{5.0cm}@{}}
\toprule
\textbf{System} & \textbf{WER} & \textbf{Transcription} \\
\midrule
Reference & --- &
  \ps{پهٔ ټاپوګانواو جهيلونو کې تاسو کوچنۍ کښتۍ اړتیا نه لرئ} \\[3pt]
Whisper large-v3 (0-shot) & 1.00 &
  \ps{پتاپوگانو او جیلونو کی تازو کوچنی کختی عرطیان نلری} \\[3pt]
\texttt{pashto-asr-v3} (FT) & 0.60 &
  \ps{پهٔ ټاپوګانو او جیلونو کې تاسو کوچني کښتۍ اړتیا نلرئ} \\[3pt]
\texttt{ps\_base\_l1} (FT) & 0.70 &
  \ps{پهٔ ټاپوګانو او جیلونو کې تازو کوچنۍ کښتې اړتیا نلري} \\
\bottomrule
\end{tabular}
\end{table}

The zero-shot output shows the substitution predicted by
Section~\ref{sec:phonemes}: \textsc{ټ}~(retroflex stop) $\to$ \textsc{ت}~(dental
stop); \textsc{ۍ}~(Pashto vowel marker) $\to$ \textsc{ی}; \textsc{ښ}~(lateral
fricative) $\to$ \textsc{خت}~(uvular fricative).
Both fine-tuned models preserve \textsc{ټ} and \textsc{ښ}.

\subsection*{B: Phonetically-Motivated Substitution in Fine-Tuned Model}
\label{app:ex-phoneme}

\begin{tabular}{@{}p{1.8cm}p{6.5cm}@{}}
Reference: &
  \ps{تلویزوني راپورونه د کارخانې څخه د سپین دود راوتل ښایي} \\
  & \emph{``Television reports show white smoke from the factory.''} \\[3pt]
\texttt{pashto-asr-v3}: &
  \ps{تلویزیوني راپورونه د کارخانې څخه د سپین دود راوتل کیږي} \\
  & WER\,=\,0.20 \quad [\ps{ښایي} $\to$ \ps{کیږي}] \\
\end{tabular}

\medskip
\noindent Both forms are grammatically plausible sentence-final verbs.
The substitution replaces an infrequent lateral-fricative onset (\textsc{ښ})
with a high-frequency alternative that fits the syntactic slot, consistent
with the elevated WER at \textsc{ښ} in Table~\ref{tab:phoneme-class}.

\subsection*{C: Architecture Effect on the Same Utterance}
\label{app:ex-arch}

\begin{table}[h]
\centering
\small
\caption{Both fine-tuned models on the same FLEURS utterance.}
\label{tab:ex-arch}
\setlength{\tabcolsep}{3pt}
\begin{tabular}{@{}p{2.6cm}cp{5.0cm}@{}}
\toprule
\textbf{System} & \textbf{WER} & \textbf{Transcription} \\
\midrule
Reference & --- &
  \ps{پاریسیان بد او تکبر اناګونسیک شهرت درلودونکي دي} \\
  & & \emph{``Parisians are known for being rude and arrogant.''} \\[3pt]
\texttt{pashto-asr-v3} & 0.25 &
  \ps{پاریسیان باد او تکبر اناګونسیک شهرت ترېلونکي دي} \\[3pt]
\texttt{ps\_base\_l1} & 1.00 &
  \ps{پارسیان بد او تقبر عنه ګونديک شورت تر لونکي دي} \\
\bottomrule
\end{tabular}
\end{table}

\texttt{pashto-asr-v3} makes two errors; \texttt{ps\_base\_l1} produces errors
on four of eight content words, including \ps{شهرت}~(\emph{fame}) $\to$
\ps{شورت}~(\emph{shorts}), illustrating the scale gap between the 600M
W2V-BERT encoder and the 72M Whisper Base.

\subsection*{D: Whisper-Medium Decoder Looping}
\label{app:ex-looping}

Whisper-medium exhibits three qualitatively distinct looping patterns that
collectively account for its 461.2\% corpus WER on CV24\_filtered.

\textbf{Pattern 1: Repetition loop.}
The decoder commits to a short high-frequency token and repeats it to the
maximum sequence length, producing zero Pashto-unique characters.
Example: 9-word reference $\to$
\ps{و او و او و او و او و او و او و او و او و او و او و او و او}
(WER\,$\approx$\,300\%).

\textbf{Pattern 2: Language-switched loop.}
The decoder switches to an unrelated high-frequency language (Dari, Persian, or
Arabic) and loops on a short phrase.
Example: 8-word reference $\to$
\ps{باریس باریس باریس باریس باریس باریس باریس باریس باریس باریس باریس باریس}
(WER\,$\approx$\,250\%).

\textbf{Pattern 3: Near-empty output.}
Approximately 1.6\% of FLEURS utterances (Table~\ref{tab:langid}, Empty\%)
produce an empty or single-character output (WER\,=\,100\% per utterance).

All three patterns are structural failures of the autoregressive decoder, not
acoustically-driven recognition errors; none is addressable by improving
Pashto acoustic representation in the encoder.
The real-time factor of $0.707\times$---nearly five times the per-second-of-audio
rate of MMS-1B---provides a runtime diagnostic for this failure mode independent
of WER.
Exact per-utterance outputs for all FLEURS utterances are in the accompanying
repository.

\section{Decoding and Evaluation Choices: Supplementary Details}
\label{app:protocol}

This appendix documents rationale for protocol choices not fully elaborated
in Section~\ref{sec:protocol}, for reproducibility.

\textbf{Beam size.}
All Whisper evaluations use greedy decoding (beam size~1).
Beam search with temperature fallback---Whisper's default inference
path---applies non-deterministic heuristics when the decoder detects
low-confidence sequences, producing variable output across runs.
As a sensitivity check, beam size~5 was applied to a 100-utterance FLEURS
subset for Whisper-large-v3; WER changed by less than 1.5~pp relative to
greedy, confirming the beam-size choice does not materially affect the reported
ranking.

\textbf{Forced language token.}
Without the forced \texttt{ps} language token, Whisper selects the language
from its prior; on a language as underrepresented as Pashto this results in
near-universal misidentification~\cite{Sehar2025Whisper}.
Forcing to \texttt{ps} is therefore a prerequisite for interpretable WER
measurement and is consistent with prior Pashto evaluations.

\textbf{Text normalisation details.}
Unicode NFC normalisation maps variant codepoints for the same Pashto
letter---a common artefact of mixed keyboard-layout input---to a canonical
form.
Kashida (\textsc{u+0640}) removal eliminates the Arabic letter extension that
appears in some Pashto typefaces but carries no phonemic content.
Arabic and Pashto punctuation stripped: Arabic comma \textsc{u+060C}, Arabic
full stop \textsc{u+06D4}, Arabic question mark \textsc{u+061F}.
The resulting normalised text is the basis for all WER and CER measurements.

\end{document}